\documentclass[10pt,journal,compsoc]{IEEEtran}
\usepackage{color}
\usepackage{amsmath}
\usepackage{algorithm}
\usepackage{amssymb}
\usepackage{algorithmic}
\usepackage{array}
\usepackage{stfloats}
\usepackage{url}
\usepackage{ragged2e}
\usepackage{cite}
\usepackage{amssymb}
\usepackage{xcolor}
\usepackage{bm}
\usepackage{setspace}
\usepackage{bigstrut}

\usepackage[utf8]{inputenc} 
\usepackage[T1]{fontenc}    
\usepackage{hyperref}       
\usepackage{url}            
\usepackage{booktabs}       
\usepackage{amsfonts}       
\usepackage{nicefrac}       
\usepackage{microtype}      
\usepackage{xcolor}         
\usepackage{booktabs}
\usepackage{graphicx}
\usepackage{multirow}
\usepackage{wrapfig}
\usepackage{makecell}

\ifCLASSOPTIONcompsoc
  \usepackage[caption=false,font=footnotesize,labelfont=sf,textfont=sf]{subfig}
\else
  \usepackage[caption=false,font=footnotesize]{subfig}
\fi

\makeatletter
\def\hlinew#1{%
  \noalign{\ifnum0=`}\fi\hrule \@height #1 \futurelet
   \reserved@a\@xhline}
\makeatother%

\definecolor{orange}{RGB}{255,127,0}

\begin{document}

\title{Learnable Distribution Calibration for Few-Shot Class-Incremental Learning}

\author{Binghao Liu, Boyu Yang, Lingxi Xie, Ren Wang, Qi Tian, Qixiang~Ye %
\IEEEcompsocitemizethanks{\IEEEcompsocthanksitem B. Liu, B. Yang and Q. Ye are with the School of Electronic, Electrical and Communication Engineering, University of Chinese Academy of Sciences, Beijing, 101408, China. E-mail: \{liubinghao18, yangboyu18\}@mails.ucas.ac.cn, qxye@ucas.ac.cn. Qixiang Ye is the corresponding author. \protect

\IEEEcompsocthanksitem L. Xie and Qi Tian are with Huawei Inc., China. Email: \{198808xc@gmail.com, tian.qi1@huawei.com\}. \protect
}
}
\IEEEtitleabstractindextext{%
\begin{abstract}
\justifying
Few-shot class-incremental learning (FSCIL) faces challenges of memorizing old class distributions and estimating new class distributions given few training samples.
In this study, we propose a learnable distribution calibration (LDC) approach, with the aim to systematically solve these two challenges using a unified framework. 
LDC is built upon a parameterized calibration unit (PCU), which initializes biased distributions for all classes based on classifier vectors (memory-free) and a single covariance matrix. The covariance matrix is shared by all classes, so that the memory costs are fixed.
During base training, PCU is endowed with the ability to calibrate biased distributions by recurrently updating sampled features under the supervision of real distributions.
During incremental learning, PCU recovers distributions for old classes to avoid `forgetting', as well as estimating distributions and augmenting samples for new classes to alleviate `over-fitting' caused by the biased distributions of few-shot samples.
LDC is theoretically plausible by formatting a variational inference procedure.
It improves FSCIL's flexibility as the training procedure requires no class similarity priori.
Experiments on CUB200, CIFAR100, and mini-ImageNet datasets show that LDC outperforms the state-of-the-arts by 4.64\%, 1.98\%, and 3.97\%, respectively. LDC's effectiveness is also validated on few-shot learning scenarios. 
\end{abstract}
}

\maketitle

\section{Introduction}

\IEEEPARstart{G}{reat} progress has been made in visual recognition, which can be attributed to advanced learning mechanisms and large-scale datasets with adequate supervision. However, machine learning remains incomparable to cognitive learning, which obtains high-precision recognition based on few supervisions and can generalize this capability to novel classes~\cite{TOPIC2020}.
To study this topic, the community starts paying attentions to few-shot class-incremental learning (FSCIL), a learning paradigm inspired by cognitive learning~\cite{Cognitive1968}. Given base classes with sufficient training data and novel classes with few samples, FSCIL is able to construct a representative model using old classes and continually adapt the model to new classes. 

\begin{figure}[t]
\begin{center}
\includegraphics[width=1\linewidth]{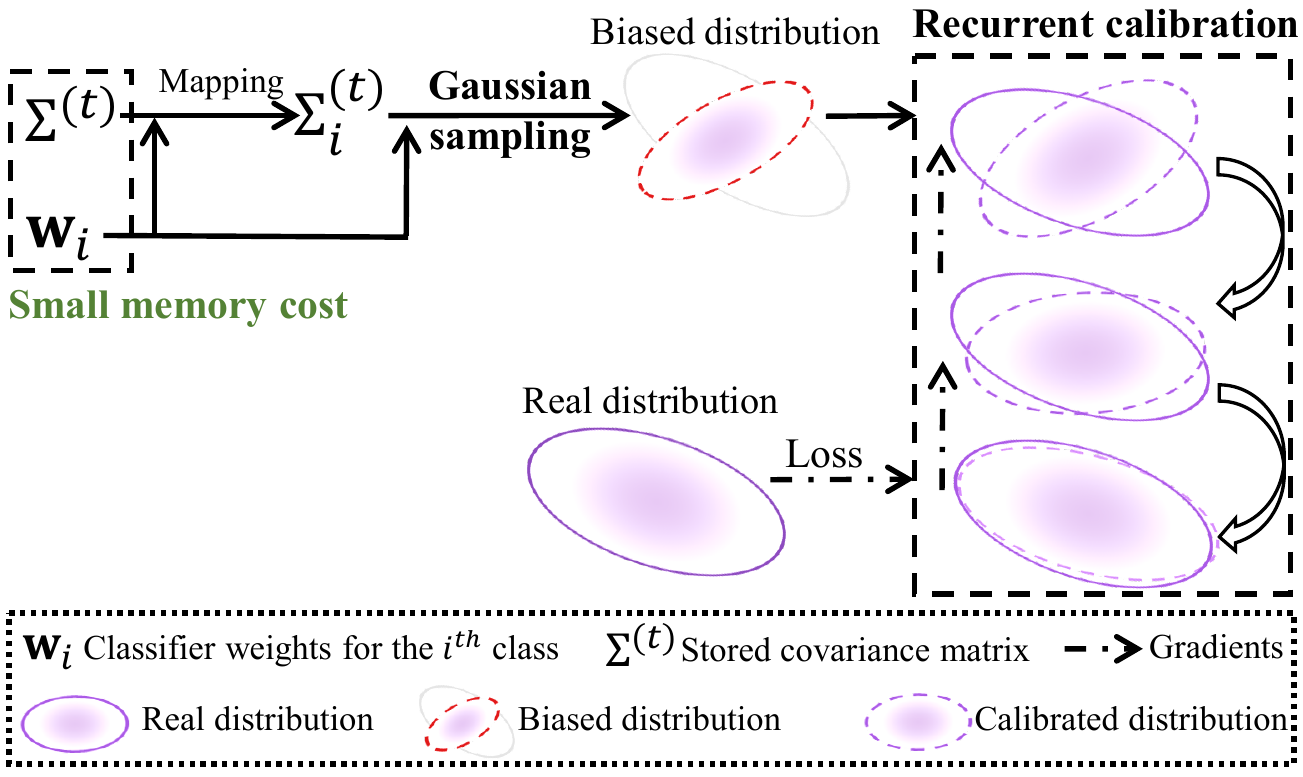} 
\end{center}
\caption{LDC calibrates biased distributions by recurrently updating sampled features under the supervision of real distributions. During few-shot class-incremental learning (FSCIL), it recovers distributions for old classes to avoid `forgetting', as well as estimating distributions and augmenting samples for new classes to alleviate `over-fitting' caused by the biased distributions of few-shot samples.}
\label{fig:LDC}
\end{figure}

\begin{figure*}[t]
\centering
\includegraphics[width=1\linewidth]{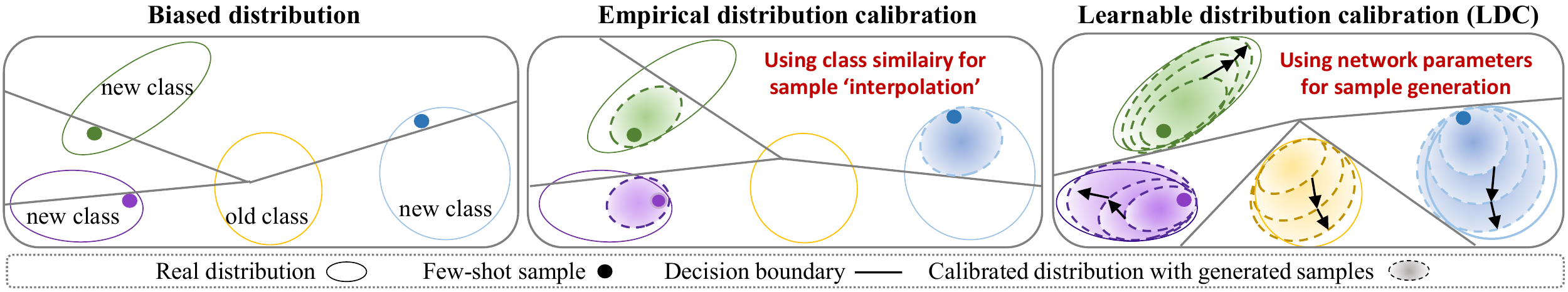}
\caption{Free-Lunch~\cite{FreeLunch2021} stores covariance matrices and mean vectors to ``interpolate" samples and solve the biased distribution issue. The required memory costs linearly increase with the class number.  In contrast, our LDC learns to generate samples using parameterized calibration units (PCU), which not only has a fixed memory overhead but also gets rid of the dependence on similarity priori. (Best viewed in color)
}
\label{fig:motivation}
\end{figure*}

However, FSCIL faces challenges that are beyond conventional learning paradigms. On the one hand, the model requires to memorize old classes and must avoid `forgetting'. On the other hand, the model needs to avoid `over-fitting' when facing biased distributions caused by few-shot samples. To alleviate `forgetting', data distillation~\cite{LwF2018,IncrementalDet2017} memorizes old class distributions by introducing regularization loss in incremental learning. However, it is challenged by biased distributions upon few-shot training samples. The Free-Lunch method~\cite{FreeLunch2021} transfers statistics from classes with sufficient samples to few-sample classes. Nevertheless, it suffers enormous memory costs when required to store covariance matrices for all base classes. Furthermore, Free-Lunch calibrates class distributions using class similarity calculated by few-shot samples, so that the calibration results are likely inaccurate as few-shot samples suffer bias.

In this study, we propose learnable distribution calibration (LDC) for FSCIL, Fig.~\ref{fig:LDC}. The purpose is to recover old class distributions and calibrate biased new class distributions in a uniform framework. Compared with the conventional method~\cite{FreeLunch2021}, LDC is differential and requires a low memory cost. It learns to calibrate distributions using a single stored covariance matrix shared by all classes. This enables getting rid of dependence on similar base classes, improving FSCIL's accuracy and flexibility, Fig.~\ref{fig:motivation}.

LDC roots on a parameterized calibration unit (PCU), which initializes feature distribution for each class using a Gaussian sampler, defined according to a mean vector and a stored covariance matrix, to generate a set of feature samples. The mean vector is approximated using the classifier vector, which is regarded as a class prototype~\cite{EvolvedClassifiers2021}. The stored covariance matrix is the mean of covariance matrices about all old classes. A learnable mapping function, which takes the stored mean covariance matrix and mean vector as inputs, is proposed to generate the covariance matrix for each class. Using base classes, PCU is trained to calibrate biased distributions by recurrently updating sampled features under the supervision of real sample features. 
The Gaussian sampler generates sufficient feature samples during incremental learning, forming biased distributions for old and new classes. PCU recurrently updates the generated feature samples, so that old class distributions are recovered and new class distributions are calibrated. We conduct sophisticated experiments on common datasets, outlier cases, and few-shot learning tasks to verify the effectiveness and generalization ability of our method.

The contributions of this study are as follows:
\begin{itemize}
    \item We propose learnable distribution calibration (LDC), with the aim to recover old class distributions and calibrate new class distributions from a single stored covariance matrix, solving catastrophically forgetting and over-fitting for FSCIL in a uniform framework.

    \item We design a parameterized calibration unit (PCU), providing a recurrent fashion for distribution calibration without using class similarity priori, improving accuracy with a fixed memory cost.

    \item LDC achieves new state-of-the-art performance on the challenging FSCIL task. It also generalizes smoothly to regular few-shot learning problems. 
\end{itemize}

\section{Related Work}

\textbf{Few-shot Learning.}
Few-shot learning trains a model to classify unseen (novel) classes with only a few annotated samples. Few-shot learning methods can be broadly categorized as either metric learning, meta learning, or data augmentation. Metric learning~\cite{MatchingNet2016,PrototypicalNet2017,RelationNet2018,DeepEMD2020,Harmonic2021, PST2021} trains two-branch networks to determine categories of the query images by comparing few-shot training images and query (test) images. Meta learning~\cite{MAML2017,MetaNeural2020,Meta-Transfer2019} pursues fast adaptation of models to new categories with few training images for optimization. Data augmentation~\cite{SaliencyHallucination2019,AdversarialHallucination2020,BoundaryAdversarial2020, FreeLunch2021} generates rich examples to rectify biased sample distributions. 

To conquer catastrophically forgetting, meta-learning~\cite{AttentionAttractor2019,XtarNet2020} and feature alignment methods~\cite{DiscriminantAlignment2020} have been explored to regularize new class training. The dynamic few-shot learning approach~\cite{DynamicFewshot2018} redesigns the classifier of a ConvNet model as the cosine similarity function between feature representations and classification weight vectors, which leads to feature representations that generalize better on ``unseen” categories. However, it doesn't involve distribution bias caused by few-shot samples, which is the focus of this study. 

\textbf{Incremental Learning.} 
According to the avaliability of task IDs, methods can be classified as either task- or class-incremental learning~\cite{IncrementalSurvey2020}. Incremental learning methods can be further grouped as either rehearsal, regularization, or architecture configuration methods. Rehearsal methods~\cite{iCaRL2017,RWalk2018,LargeScaleIL2019,DeepReplay2017,CAN2019, SPR2021,AlwaysBeDreaming2021} recall exemplars, which are stored from a previous session, to prevent forgetting. Regularization methods~\cite{LwF2018,LwM2019,PathInt2017,GradCAM2017,RWalk2018,NCM2019,DistillingCausalEffect2021,Co2L2021} introduce special loss functions that utilize knowledge or parameter distillation to constrain network learning. Architecture configuration methods~\cite{PackNet2018,HardAttention2018,DEN2018,Piggyback2018} choose parts of network parameters by leveraging hard attention~\cite{HardAttention2018}, pruning mechanisms~\cite{PackNet2018} or dynamic expansion network~\cite{DEN2018} to alleviate model drift.

When task IDs are not accessible during inference, the problem evolves into class-incremental learning~\cite{iCaRL2017}. The primary challenge is catastrophic forgetting, which has been elaborated by using data distillation~\cite{LwF2018,IncrementalDet2017}, memory mechanisms~\cite{LwM2019,Mnemonics2020,IL2M2019}, and transfer learning~\cite{Transfer19}. Exemplars are selected based upon class feature distribution~\cite{iCaRL2017}, or produced by generative adversarial networks~\cite{CAN2019}. 

\textbf{Few-shot Class-Incremental Learning.} This task manages to train a representation model using base classes and continually adapts the model to new classes. FSCIL therefore amalgamates the challenges of catastrophic forgetting caused by incremental learning and over-fitting caused by biased few-shot samples. The neural gas method~\cite{TOPIC2020} considered these challenges by constructing and preserving feature topology. 
Continually evolving prototypes~\cite{Self-Promoted2021,EvolvedClassifiers2021} learns new classes by optimizing class margins. The mixture sub-space approach~\cite{SynthesizedFeature2021} synthesizes features in the sub-spaces for incremental classes by using a variational auto-encoder.

Despite substantial progress, the problem of solving the forgetting and over-fitting issues in a uniform framework remains. The distribution calibration method~\cite{FreeLunch2021} initiated the idea to solve the over-fitting issue, but has difficult to handle FSCIL as the memory cost increase with the class number. 

\begin{figure*}[t]
\centering
\includegraphics[width=1.0\linewidth]{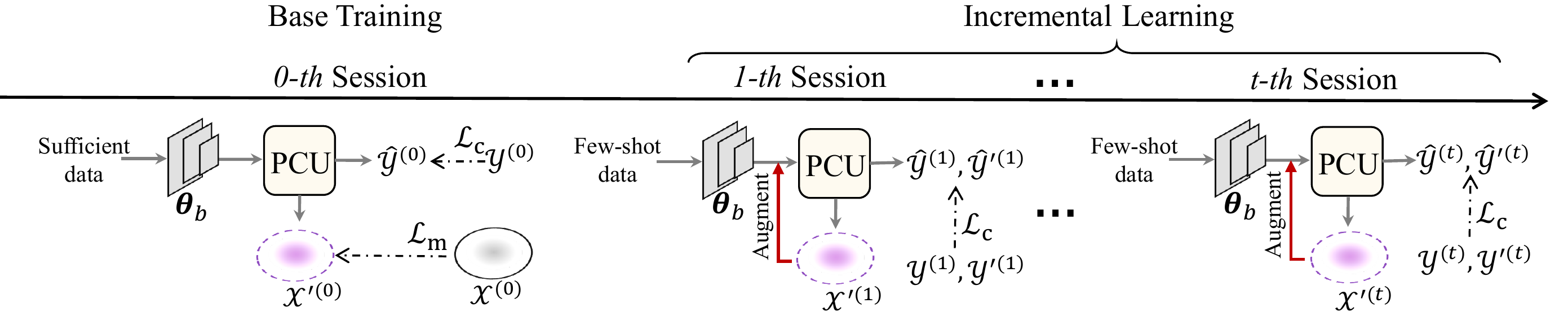}
\caption{Flowchart of learnable distribution calibration (LDC) for few-shot class-incremental learning (FSCIL).}
\label{fig:flowchart}
\end{figure*}

\section{The Proposed Approach}

\subsection{Overview}
\label{section:overview}
FSCIL consists of base training and incremental learning stages, Fig.~\ref{fig:flowchart}. During base training, it learns a representation model upon base classes ($\mathcal{C}^{(0)}$) with sufficient examples. During incremental learning, it generalizes the developed model to new classes using only a few samples. The incremental datasets are denoted as $\{\mathcal{D}^{(t)}, t=0,1,2,...\}$, where $\mathcal{D}^{(t)}$ is the data of classes $\mathcal{C}^{(t)}$ for the $t$-th session. For $t_1 \neq t_2$, we have $\mathcal{C}^{(t_1)} \cap \mathcal{C}^{(t_2)}=\varnothing$. In the $t$-th session, only the training data $\mathcal{D}^{(t)}$ for the class set $\mathcal{C}^{(t)}$ is available, while all old and new classes $\{\mathcal{C}^{(0)}, ..., \mathcal{C}^{(t)}\}$ require to be tested. That said, old classes $\{\mathcal{C}^{(0)},\cdots,\mathcal{C}^{(t-1)}\}$ shall not be forgotten while the new concepts shall be learned with few samples.

\begin{figure}[t]
\begin{center}
\includegraphics[width=0.85\linewidth]{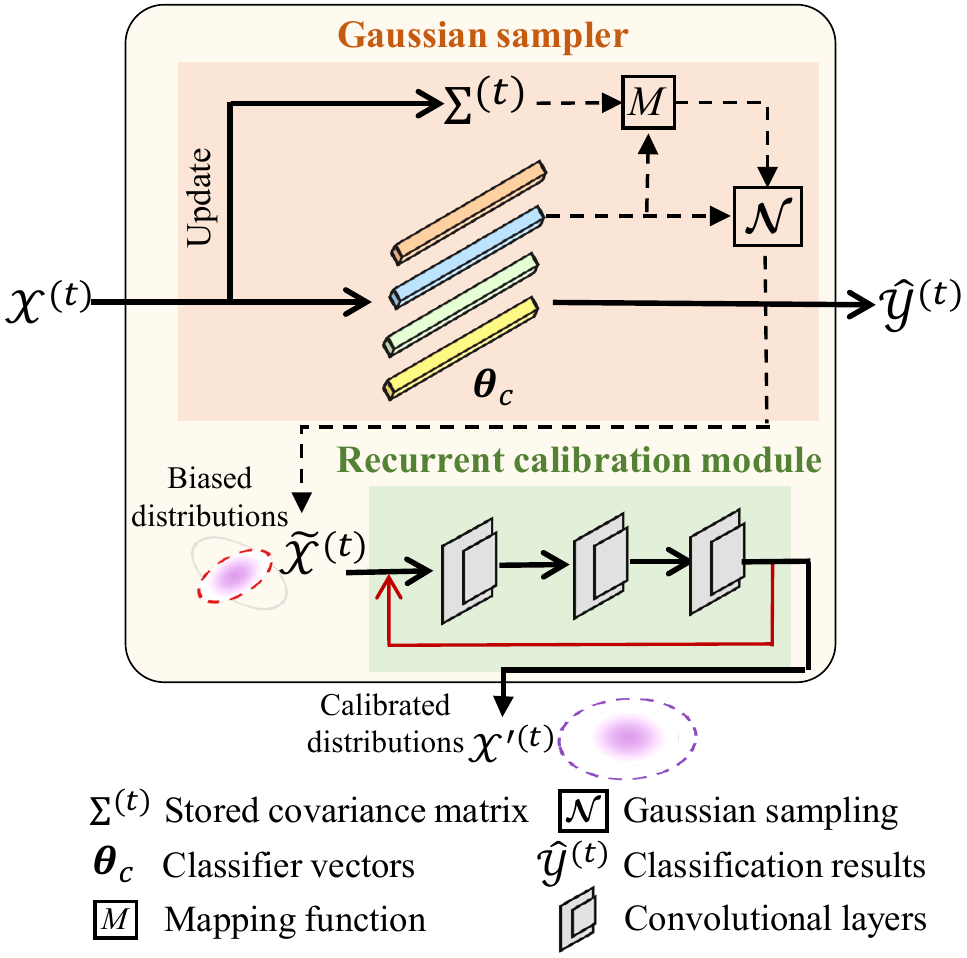} 
\end{center}
\caption{Parameterized calibration unit.}
\label{fig:pcu}
\end{figure}

Fig.~\ref{fig:flowchart} shows LDC's flowchart, where the network consists of a feature extractor ($\boldsymbol{\theta}_b$) and a PCU. The PCU contains a Gaussian sampler that initiates biased distributions, a learnable mapping function ($\boldsymbol{\theta}_m$) that generates the covariance matrix for each class, a recurrent calibration module ($\boldsymbol{\theta}_r$) that manipulates biased distributions, and a classifier ($\boldsymbol{\theta}_c$), Fig.~\ref{fig:pcu}. During base training, images are encoded to feature samples $\mathcal{X}^{(0)}$ by a feature extractor ($\boldsymbol{\theta}_b$), which are then fed into the PCU for classification prediction ($\mathcal{\hat{Y}}^{(0)}$). The feature extractor is trained under the supervision of ground-truth labels $\mathcal{Y}^{(0)}$. At the same time, PCU learns to generate feature samples $\mathcal{X}^{\prime(0)}$ (calibrated distributions)
under the supervision of the real feature samples $\mathcal{X}^{(0)}$ (which represent real distributions). The base training procedure is formulated as:
\begin{equation}
\begin{split}
    &\mathop{\arg\min}_{\boldsymbol{\theta}_b,\boldsymbol{\theta}_c,\boldsymbol{\theta}_m,\boldsymbol{\theta}_r} 
    \mathcal{L}_{\mathrm{c}}(\mathcal{\hat{Y}}^{(0)}, \mathcal{Y}^{(0)}; \boldsymbol{\theta}_b, \boldsymbol{\theta}_c) +\mathcal{L}_{\mathrm{m}}(\mathcal{X}^{\prime(0)}, \mathcal{X}^{(0)}; \boldsymbol{\theta}_m, \boldsymbol{\theta}_r), 
    \\&s.t. ~\hat{\mathcal{Y}}^{(0)}=g(\mathcal{X}^{(0)};{\boldsymbol{\theta}_c}),  
\end{split}
  \label{eq:loss_base}
\end{equation}
where $\mathcal{L}_{\mathrm{c}}$ and $\mathcal{L}_{\mathrm{m}}$ respectively denote image classification loss and distribution matching loss. $\mathcal{L}_{\mathrm{c}}$ calculates the cross entropy between predictions and labels, and $\mathcal{L}_{\mathrm{m}}$ measures the distance between calibrated and real distributions. $g(\cdot)$ denotes the classification procedure.

In the $t$-th incremental session, the PCU trained upon base classes is utilized to recover old class distributions and generalized to new classes for distribution calibration. Old class samples (which is unavailable) are recovered in two steps: (1) Initializing feature samples using the Gaussian sampler; (2) Recurrently calibrating the feature samples so that their distributions approach the real ones. These two steps are also carried out for the new classes to augment training samples and calibrate the biased distributions. Few-shot samples are used to update classifier vectors and the stored covariance matrix, which are used to perform Gaussian sampling. The sampled features are then fed to the recurrent calibration module to generate calibrated feature samples for model training. The incremental learning procedure is formulated as:
\begin{equation}
\begin{split}
    &\mathop{\arg\min}_{\boldsymbol{\theta}_c} \mathcal{L}_{\mathrm{c}}(\hat{\mathcal{Y}}^{(t)}, \mathcal{Y}^{(t)}; \boldsymbol{\theta}_c)+ \mathcal{L}_{\mathrm{c}}(\hat{\mathcal{Y}}^{\prime(t)}, \mathcal{Y}^{\prime(t)}; \boldsymbol{\theta}_c), 
    \\
    &s.t.~\hat{\mathcal{Y}}^{(t)}=g(\mathcal{X}^{(t)};{\boldsymbol{\theta}_c}),
                  ~\hat{\mathcal{Y}}'^{(t)}=g(\mathcal{X}'^{(t)};{\boldsymbol{\theta}_c}),
 \end{split} 
 \label{eq:loss_inc}
\end{equation}
where $\mathcal{X}^{(t)}$ denotes few-shot samples for the $t$-th session and $\mathcal{X}'^{(t)}$ denotes calibrated feature samples (calibrated distributions). $\hat{\mathcal{Y}}^{(t)}$ and $\mathcal{Y}^{(t)}$ denote the predicted and ground-truth labels of few-shot samples. $\hat{\mathcal{Y}}^{\prime(t)}$ and $\mathcal{Y}^{\prime (t)}$ denote the predicted and ground-truth labels of the calibrated samples, respectively. In order to handle new classes $\mathcal{C}^{(t)}$, $N^{(t)}-N^{(t-1)}$ classifier vectors are added to the classification layer as $\boldsymbol{\theta}_c = [\mathbf{w}_0,...,\mathbf{ w}_{N^{(t-1)}},...,\mathbf{w}_{N^{(t)}}]^\intercal$, where $N^{(t)}$ denotes the number of seen classes until the $t$-th session. The network is updated by optimizing Eq.~\ref{eq:loss_inc}.

\subsection{Parameterized Calibration Unit}
\label{section:PCU}
PCU consists of a Gaussian sampler and a recurrent calibration module, both of which are parameterized, Fig.~\ref{fig:pcu}.

\textbf{Gaussian Sampler.} It is defined by multiplexing classifier vectors ($\boldsymbol{\theta}_c$) as mean vectors and storing a shared covariance matrix ($\boldsymbol{\Sigma}^{(t)}$) which is the mean covariance matrix for all classes. The classifier is calculates the cosine distance between sample features and classifier vectors for classification~\cite{closerfewshot}. Classifier vectors are initialized by the mean feature vectors of the corresponding classes. The $i$-th classifier vector ($\mathbf{w}_i$) can be approximated as the prototype of the $i$-th class. The covariance matrix ($\boldsymbol{\Sigma}^{(t)}$) is dynamically updated as $\boldsymbol{\Sigma}^{(t)} = \boldsymbol{\Sigma}^{(t-1)}\cdot\frac{N^{(t-1)}}{N^{(t)}} + \mathrm{Cov}(\mathcal{X}^{(t)})\cdot\frac{N^{(t)}-N^{(t-1)}}{N^{(t)}}$. $\mathrm{Cov}(\cdot)$ calculates the covariance matrix. A learnable mapping function ($M$) is used to generate the covariance matrix for the $i$-th class, as $\boldsymbol{\Sigma}^{(t)}_i=M(\mathbf{w}_i, \boldsymbol{\Sigma}^{(t)}; \boldsymbol{\theta}_m)$, which is realized by an expanding-concatenating operation and a Conv-ReLU-Conv stack. The $i$-th covariance matrix and classifier vector are fed to the sampler to generate a biased distribution ($\tilde{\mathcal{X}}_i^{(t)}$) for the $i$-th class. The sampling process is expressed as $\tilde{\mathcal{X}}_i^{(t)} = \{x |x \sim \mathcal{N}(\mathbf{w}_i,\boldsymbol{\Sigma}^{(t)}_i)\}$, where $x$ are the features sampled from the Gaussian distribution $\mathcal{N}(\mathbf{w}_i,\boldsymbol{\Sigma}^{(t)}_i)$. For the $t$-th session, we generate the biased distributions for all seen classes, which are calculated as $\tilde{\mathcal{X}}^{(t)} = \{ \tilde{\mathcal{X}}_0^{(t)}, \tilde{\mathcal{X}}_1^{(t)}, \cdots, \tilde{\mathcal{X}}_{N^{(t)}}^{(t)} \}$.

\textbf{Recurrent Calibration module.} This is defined using convolutional layers, parameterized with $\boldsymbol{\theta}_r$, Fig.~\ref{fig:pcu}. For the $t$-th session, the recurrent calibration process is expressed as:
\begin{equation}
\begin{split}
&\mathcal{X}'^{(t)} = f(\Tilde{\mathcal{X}}^{(t)};{\boldsymbol{\theta}_r}),\\
&~\Tilde{\mathcal{X}}^{(t)}\leftarrow{\mathcal{X}'}^{(t)}.
\end{split}
\label{eq:recurrent}
\end{equation}
During recurrent calibration, the objective function is built upon previous optimization and prediction. This makes it possible to deconstruct a complex objective function ($e.g.$, the distribution matching loss) to series of simpler objective functions which are easier to realize.

\textbf{Distribution Matching Loss.} 
During base training, the shared covariance matrix and classifier vectors are used to sample biased distributions. Then those biased distributions are calibrated to obtain the calibrated distributions $\mathcal{X}'^{(0)}$. Given the real distributions $\mathcal{X}^{(0)}$, the loss is defined as: 
\begin{equation}  
   \mathcal{L}_{\mathrm{m}}(\mathcal{X}'^{(0)},\mathcal{X}^{(0)}; \boldsymbol{\theta}_m, \boldsymbol{\theta}_r)  = \mathrm{KL}(\mathcal{X}'^{(0)}||\mathcal{X}^{(0)}),
   \label{eq:loss_matching}  
\end{equation}  
where $\mathrm{KL}(\cdot)$ is defined as the KL divergence~\cite{KL}.
By minimizing the distribution matching loss $\mathcal{L}_{\mathrm{m}}$, PCU's parameters are updated so that the calibrated distributions align with real distributions. PCU is generalized to new classes for calibration during incremental learning.

\subsection{Theoretical Analysis}
\label{section:analysis}
We managed to prove that recurrent sample calibration is theoretically plausible, by formulating it as a variational inference procedure during training~\cite{staticml}. Given real distributions $\mathcal{X}^{(t)}$, biased distributions $\tilde{\mathcal{X}}^{(t)}$ and the parameters $\boldsymbol{\theta}_r$ for recurrent distribution calibration. $(\boldsymbol{\theta}_r,\tilde{\mathcal{X}}^{(t)})$ defines latent variable $\mathbf{z}$. The optimization of $\mathbf{z}$ is equivalent to finding a mean-field distribution~\cite{mean_field} $q^*(\mathbf{z})$ which is  to approximate the distribution $p(\mathbf{z}|\mathcal{X}^{(t)})$, which is formulated as:
\begin{equation}
    q^*(\mathbf{z})= \mathop{\arg\min}_{q(\mathbf{z})\in \mathcal{Q}} \mathrm{KL}(q(\mathbf{z})||p(\mathbf{z}|\mathcal{X}^{(t)})), 
    \label{eq:qz1}
\end{equation}
where $\mathcal{Q}$ denotes the candidate probability distributions. According to the characteristics of mean-field distribution, $q(\mathbf{z})=\prod_{m=1}^M q_m(\mathbf{z}_m)$ where $q_m(\mathbf{z}_m)$ is a subset of $q(\mathbf{z})$. $\mathrm{KL}$ denotes the Kullback-Leibler divergence between distributions, which is calculated by:
\begin{equation}
\begin{aligned}
    \mathrm{KL}(q(\mathbf z)||p(\mathbf z|\mathcal{X}^{(t)}))&=\mathbb{E}_q[\log q(\mathbf z)]- \mathbb{E}_q[\log p(\mathbf z|\mathcal{X}^{(t)})]\\
    &=\mathbb{E}_q[\log q(\mathbf z)]-\mathbb{E}_q[\log p(\mathcal{X}^{(t)},\mathbf z)]\\
    &+ \log p(\mathcal{X}^{(t)})\\
    &=\log p(\mathcal{X}^{(t)}) - \{\mathbb{E}_q[\log p(\mathcal{X}^{(t)},\mathbf z)]\\
    &-\mathbb{E}_q[\log q(\mathbf z)]\}.
    \label{eq:sup_kl}
\end{aligned}
\end{equation}
$\mathbb{E}$ denotes the expectation function. $\mathrm{KL}$ is non-negative, thus we have:
\begin{equation}
    \log p(\mathcal{X}^{(t)}) \geq \mathbb{E}_q[\log p(\mathcal{X}^{(t)},\mathbf z)]-\mathbb{E}_q[\log q(\mathbf z)].
    \label{eq:sup_elbo}
\end{equation}
The right side of the Eq.~\ref{eq:sup_elbo} is the lower bound of the left side, which is termed as evidence lower bound ($\mathrm{ELBO}$). We have:
\begin{equation}
\begin{aligned}
    \mathrm{KL}(q(\mathbf z)||p(\mathbf z|\mathcal{X}^{(t)}))&=\log p(\mathcal{X}^{(t)}) - \{\mathbb{E}_q[\log p(\mathcal{X}^{(t)},\mathbf z)]\\
    &-\mathbb{E}_q[\log q(\mathbf z)]\}\\
    &= \log p(\mathcal{X}^{(t)})-\mathrm{ELBO}(q(\mathbf z),\mathcal{X}^{(t)}).
    \label{eq:loglike}
\end{aligned}
\end{equation}
According to Eq.~\ref{eq:loglike}, Eq.~\ref{eq:qz1} is re-written as:
\begin{equation}
    q^*(\mathbf{z})= \mathop{\arg\min}_{q(\mathbf{z})\in \mathcal{Q}} \log p(\mathcal{X}^{(t)})-\mathrm{ELBO}(q(\mathbf{z}),\mathcal{X}^{(t)}).
    \label{eq:qz2}
\end{equation}
As $p(\mathcal{X}^{(t)})$ is independent to $q(\mathbf{z})$, Eq.~{\ref{eq:qz2}} is equivalent to:
\begin{equation}
    q^*(\mathbf{z})= \mathop{\arg\max}_{q(\mathbf{z})\in \mathcal{Q}} \mathrm{ELBO}(q(\mathbf{z}),\mathcal{X}^{(t)}),
    \label{eq:qz3}
\end{equation}
for the purpose of simplification, we use $\mathrm{ELBO}$ for $\mathrm{ELBO}(q,\mathcal{X}^{(t)})$, where $\mathrm{ELBO}$ is calculated as:
\begin{equation}
\begin{aligned}
    \mathrm{ELBO}&=\int q(\mathbf{z})\log \frac{p(\mathcal{X}^{(t)},\mathbf{z})}{q(\mathbf{z})}d\mathbf{z} \\
    &= \int \prod_{m=1}^M q_m(\mathbf{z}_m)(\log p(\mathcal{X}^{(t)},\mathbf{z}) \\
    &- \sum_{m=1}^M \log q_m(\mathbf{z}_m))d\mathbf{z}.
\end{aligned}
\label{eq:elbo1}
\end{equation}
For $\forall j \in[1,M]$,
the term $\prod_{m=1}^M q_m(\mathbf{z}_m)$ is expressed as $q_j(\mathbf{z}_j)\times \prod_{m\neq j}q_m(\mathbf{z}_m)$, and the term $\sum_{m=1}^M \log q_m(\mathbf{z}_m))$ is expressed as $q_j(\mathbf{z}_j)+\sum_{m\neq j} \log q_m(\mathbf{z}_m))$. Accordingly, Eq.~\ref{eq:elbo1} is re-written as:
\begin{equation}
\begin{aligned}
    \mathrm{ELBO}&=\int q_j(\mathbf{z}_j)(\int \prod_{m\neq j}q_m(\mathbf{z}_m)\log p(\mathcal{X}{(t)},\mathbf{z})d\mathbf{z}_m)d\mathbf{z}_j \\
    &- \int q_j(\mathbf{z}_j) \log q_j(\mathbf{z}_j)d\mathbf{z}_j + \mathrm{c}\\
    &= \int q_j(\mathbf{z}_j) \log \tilde{p}(\mathcal{X}^{(t)},\mathbf{z}_j)d\mathbf{z}_j \\
    &- \int q_j(\mathbf{z}_j) \log q_j(\mathbf{z}_j) d\mathbf{z}_j + \mathrm{c}, 
\label{eq:elbo2}
\end{aligned}
\end{equation}
where $\mathrm{c}$ denotes a constant. $\log \tilde{p}(\mathcal{X}^{(t)},\mathbf{z}_j)$ is the non-normalized distribution of $\mathbf{z}_j$, calculated using:
\begin{equation}
\begin{aligned}
    \log \tilde{p}(\mathcal{X}^{(t)},\mathbf{z}_j) &= \int \prod_{m\neq j}q_m(\mathbf{z}_m)\log p(\mathcal{X}{(t)},\mathbf{z})d\mathbf{z}_m) \\
    &= \mathbb{E}_{q(\mathbf{z}_{\backslash j})}[\log p(\mathcal{X}^{(t)},\mathbf{z})] + \mathrm{const},
    \label{eq:log}
\end{aligned}
\end{equation}
where $\mathbf{z}_{\backslash j}$ are the latent variables other than $\mathbf{z}_j$. Assuming the $\mathbf{z}_{\backslash j}$ is fixed, we optimize $q_j(\mathbf{z}_j)$ to maximize the $\mathrm{ELBO}$. According to Eqs.~\ref{eq:elbo2} and ~\ref{eq:log}, the optimal $q^{*}_j(\mathbf{z}_j)$ is proportional to the expectation of $\log p(\mathcal{X}^{(t)},\mathbf{z})$, as:
\begin{equation}
    q^{*}_j(\mathbf{z}_j) = \tilde{p}(\mathcal{X}^{(t)},\mathbf{z}_j) \propto \exp (\mathbb{E}_{q(\mathbf{z}_{\backslash j})}[\log p(\mathcal{X}^{(t)},\mathbf{z})]).
    \label{eq:propto}
\end{equation}
According to Eq.~\ref{eq:propto}, by recurrently optimizing $q^{*}_j(\mathbf{z}_j)$, which is achieved by calibrating $\tilde{\mathcal{X}}^{(t)}$ and updating $\boldsymbol{\theta}_r$, $\mathrm{ELBO}$ is maximized and converges to a local optimal solution. According to Eqs.~~\ref{eq:qz2} and ~\ref{eq:qz3}, maximizing $\mathrm{ELBO}$ is equal to minimizing $\mathrm{KL}(q(\mathbf{z})||p(\mathbf{z}|\mathcal{X}^{(t)}))$, which infers that a feasible solution for $q^*(\mathbf{z})$ could be found so that the calibrated distributions $\mathcal{X}'^{(t)}$ in Eq.~\ref{eq:recurrent} approximate real ones ${\mathcal{X}}^{(t)}$.

\section{Experiment}

\subsection{Settings}
\textbf{Dataset}. We evaluate LDC using three commonly used datasets, including CIFAR100, CUB200, and mini-ImageNet. The categories in the datasets are divided to base ones with adequate annotations and new ones with \textit{K}-shot annotated images. For the FSCIL, the proposed LDC is trained upon base classes for the first session. New classes are gradually added to train LDC in \textit{T} incremental sessions. In each incremental session, \textit{N}-way new classes are added for model training. CIFAR100 and mini-ImageNet datasets consist of 100 classes, where 60 classes are set as base classes and 40 as new classes. Each new class has 5-shot annotated images (\textit{K} = 5). New classes are then divided to eight sessions (\textit{T} = 8), each of which has five classes (\textit{N} = 5). CUB200 consists of 200 classes where 100 classes are set as base classes and the other 100 classes as new classes under the settings of \textit{K} = 5, \textit{T} = 10, \textit{N} = 10.

\setlength{\tabcolsep}{4pt}
\begin{table*}[t]
\caption{Evaluation of the proposed `Gaussian Sampler' and `Recurrent Calibration Module'.}
\label{table:PCU}
\begin{center}
\renewcommand\tabcolsep{4.0pt}
\resizebox{0.9\linewidth}{12.5mm}{
\begin{tabular}{cccccccccccccccccc}
\hline
\hline
\multirow{2}{*}{Baseline} & \multirow{2}{*}{\makecell[c]{Gaussian\\ Sampler}} & \multirow{2}{*}{\makecell[c]{Recurrent\\ Calibration}} & \multicolumn{11}{c}{Accuracy in each session (\%)}\\
& & &0 &1 & 2 & 3 & 4 & 5 & 6 & 7 & 8 & 9 & 10\\
\hline
\checkmark &  &  & 77.89 &74.01 & 69.23 & 64.85 & 62.35 & 59.11 &56.87 & 55.68 & 53.92 &52.79 & 51.86\\
 \checkmark& \checkmark&  &77.89&75.68 &71.12 & 67.31& 66.13 &60.89 & 58.73 & 57.65 & 56.08 & 55.21 & 54.58 \\ 
 \checkmark& \checkmark& \checkmark &77.89 &76.93 &74.64 &70.06 &68.88 &67.15 &64.83 &64.16 &63.03 &62.39 &61.58\\ 
\hline
\end{tabular} 
}
\end{center} 
\end{table*}

\begin{table*}[t]
\caption{Comparison of distribution matching losses. `EMD' denotes the earth mover distance.} 
\label{table:matchingloss}
\begin{center} 
\renewcommand\tabcolsep{4.0pt}
\resizebox{0.9\textwidth}{6.4mm}{
\begin{tabular}{l|ccccc}
\hline
\hline
Distribution Loss &EMD~\cite{emd} & KL divergence~\cite{KL} & JS divergence~\cite{JS} & Hellinger distance~\cite{HD} \\
\hline
Accuracy on the last session (\%) &59.21 &\bf{61.58} &58.68 & 58.32\\
\hline
\end{tabular}
}
\end{center}
\end{table*}

\textbf{Training and Evaluation.} Following ~\cite{TOPIC2020,EvolvedClassifiers2021,fact}, we adopt Resnet-18~\cite{ResNet16} as the backbone for CUB200 and mini-ImageNet, and Resnet-20 for CIFAR100. The model is optimized using the SGD algorithm. Following FACT~\cite{fact}, we use normalization, horizontal flipping, random cropping, auto augmenting and random resizing for data augmentation. For the first session, we train the network using $D^{(0)}$ upon base classes, with a batch size 128 and an initial learning rate 0.1. When $t>0$, the network is trained upon $D^{(t)}$ with new classes and the learning rate is set to 0.01. The network is trained for 100 epochs during each session. During inference period, we conduct 10 experiments using random seeds and report averaged results to eliminate the randomness of the experiments. We adopt a simple baseline method which consists of a feature extractor and a distance-based classifier. Experiments are conducted using PyTorch 1.9.0 and run on Nvidia 2080Ti GPUs.

During inference, the model trained for all $T$ sessions are evaluated by classification on all the seen class $\{\mathcal{C}^{(0)}, \mathcal{C}^{(1)},...,\mathcal{C}^{(t)}\}$ using the metric of $\mathrm{accuracy}=\frac{\mathrm{TP}+\mathrm{TN}}{\mathrm{TP+TN+FP+FN}}$, where $\mathrm{TP}$, $\mathrm{TN}$, $\mathrm{FP}$ and $\mathrm{FN}$ respectively denote the numbers of true positives, true negatives, false positives and false negatives. 
To better measure model forgetting,
`$\mathrm{PD}$' and `$\mathrm{PR}$' are introduced to evaluate the model performance drop and performance retention respectively. `$\mathrm{PD}$' is defined as $\mathrm{PD} = \mathrm{accuracy}^{(0)}-\mathrm{accuracy}^{(T)}$, where $\mathrm{accuracy}^{(0)}$ denotes the $0$-th session performance and $\mathrm{accuracy}^{(T)}$ denotes the $T$-th session performance. `$\mathrm{PR}$' is defined as $\mathrm{PR} = \mathrm{accuracy}^{(T)}/\mathrm{accuracy}^{(0)}$.

\subsection{Ablation Study}

We conduct ablation studies on CUB200 to validate PCU's modules, including the Gaussian sampler, the recurrent calibration module and the distribution matching loss.

\textbf{Gaussian Sampler.} In Table~\ref{table:PCU}, the proposed approach outperforms the baseline method by 2.72\% with `Gaussian Sampling'. The performance gain validates the plausibility of the Gaussian sampling process based on classifier weights and a single stored covariance matrix. 

\textbf{Recurrent Calibration.} With recurrent calibration, LDC further improves the baseline method and Gaussian sampler by 9.72\% and 7.00\%, respectively. By recurrently calibrating biased distributions generated with the Gaussian sampler, generated distributions approach real distributions, which augments training data and benefits model learning.

\textbf{Distribution Matching Loss.} Four kinds of matching losses defined upon earth mover distance, KL divergence, JS divergence, and Hellinger distance are compared in Table~\ref{table:matchingloss}, where one can see that KL divergence performs the best. 

\begin{figure}[t]
\begin{center}
\includegraphics[width=1.0\linewidth]{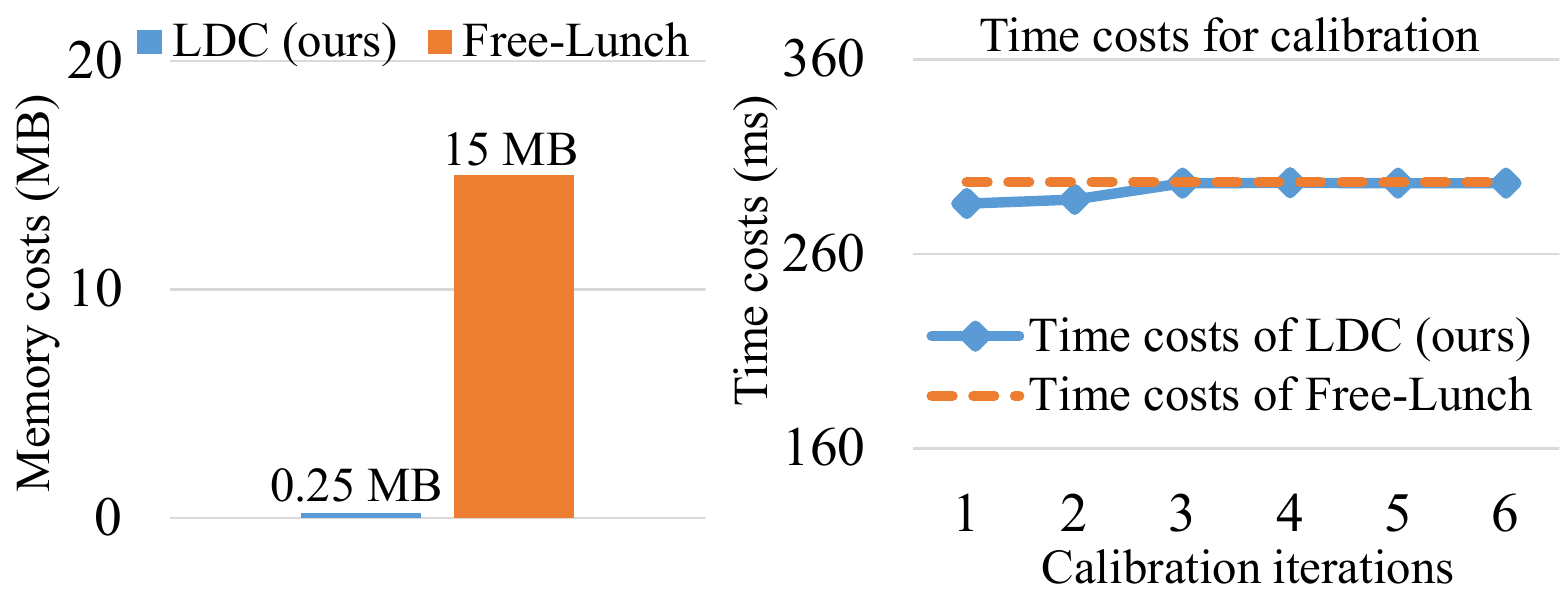} 
\end{center}
\caption{Memory and time costs on mini-ImageNet.}
\label{fig:iteration}
\end{figure}

\textbf{Memory and Time Costs.} As shown in Fig.~\ref{fig:iteration}, LDC's memory costs for covariance matrix storage is much lower than Free-Lunch's. And LDC's time costs is comparable to Free-Lunch's when the calibration iterations increase. It is attributed to the employing of convolution layers so that recurrent calibration costs negligible time.

\subsection{Model Analysis}

\begin{figure}[t]
\centering
\includegraphics[width=1.0\linewidth]{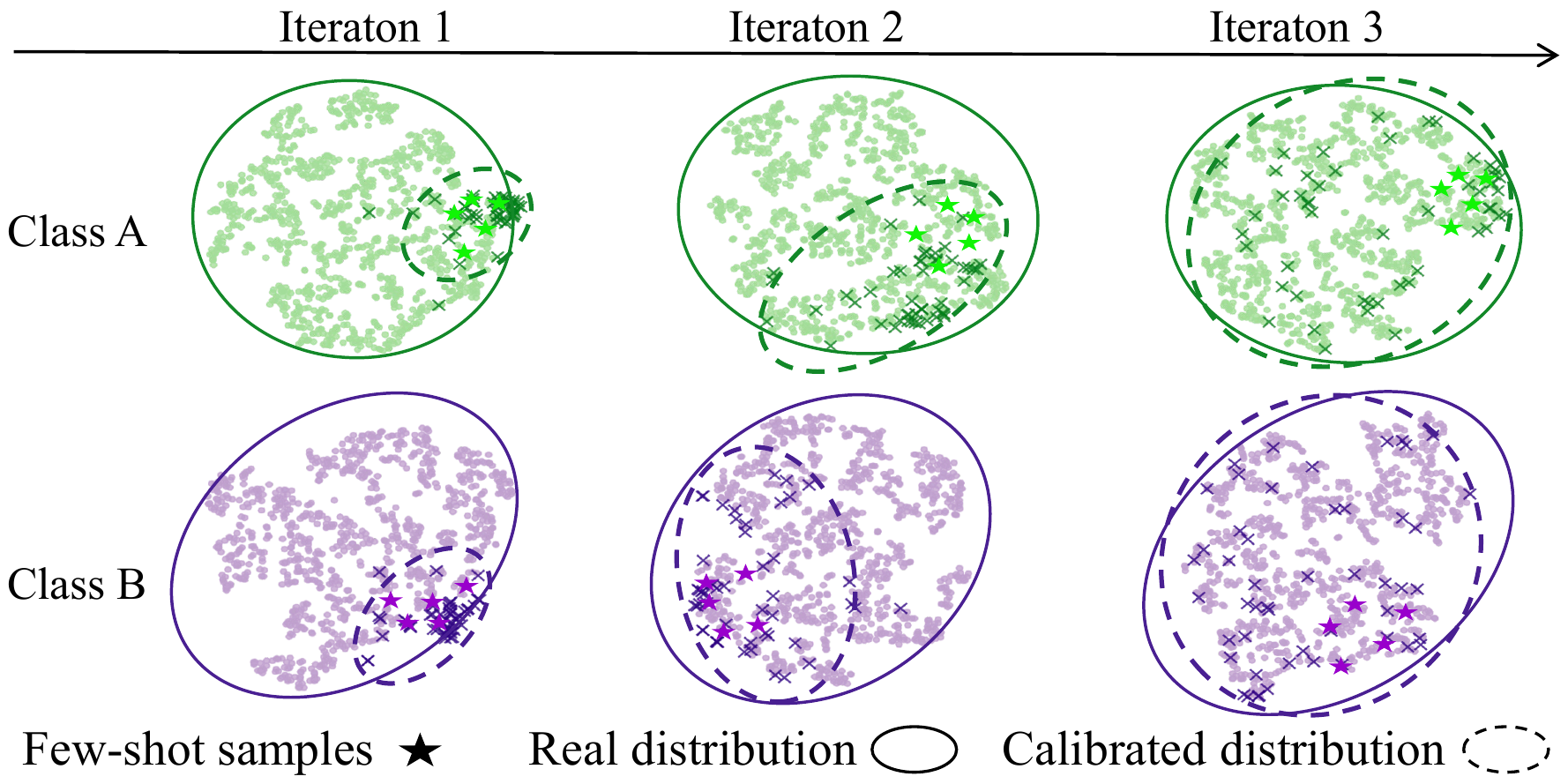}
\caption{Visualization of distribution calibration using LDC. (Best viewed in color)}
\label{fig:DC}
\end{figure}

\textbf{Distribution Calibration.} 
We utilize t-SNE to visualize real distributions and calibrated distributions generated using the proposed LDC methods to analyze the recurrent distribution calibration process better. Fig.~\ref{fig:DC} shows that the calibrated distributions generated by LDC approach real distributions when calibration proceeds. Consequently, training with calibrated distribution samples alleviates catastrophic forgetting and benefits estimations for new class distributions.

\begin{table}[h]
\caption{Comparison of regular and outlier cases.} 
\label{table:outlier}
\centering
\renewcommand\tabcolsep{4.0pt}
\resizebox{0.9\linewidth}{12mm}{
\begin{tabular}{l|l|cc|cccc}
\hline
\hline
\multicolumn{2}{c}{Method} &$\mathrm{PD} \downarrow$ & $\mathrm{PR} \uparrow$ & $\Delta \mathrm{PD}$ & $\Delta \mathrm{PR}$ \\
\hline
\multirow{2}{*}{Normal} & Free-Lunch &25.62 &65.65\% &\multirow{2}{*}{3.11} & \multirow{2}{*}{4.22\%} \\ &\bf LDC (ours) &\bf22.51 &\bf69.87\% \\
\hline
\multirow{2}{*}{Outlier} & Free-Lunch &30.07&59.68$\%$ &\multirow{2}{*}{\bf4.75} & \multirow{2}{*}{\bf6.11\%}\\ &\bf LDC (ours) &\bf 25.32 &\bf65.79$\%$\\
\hline
\end{tabular}}
\end{table}

\begin{figure}
\begin{center}
\includegraphics[width=1\linewidth]{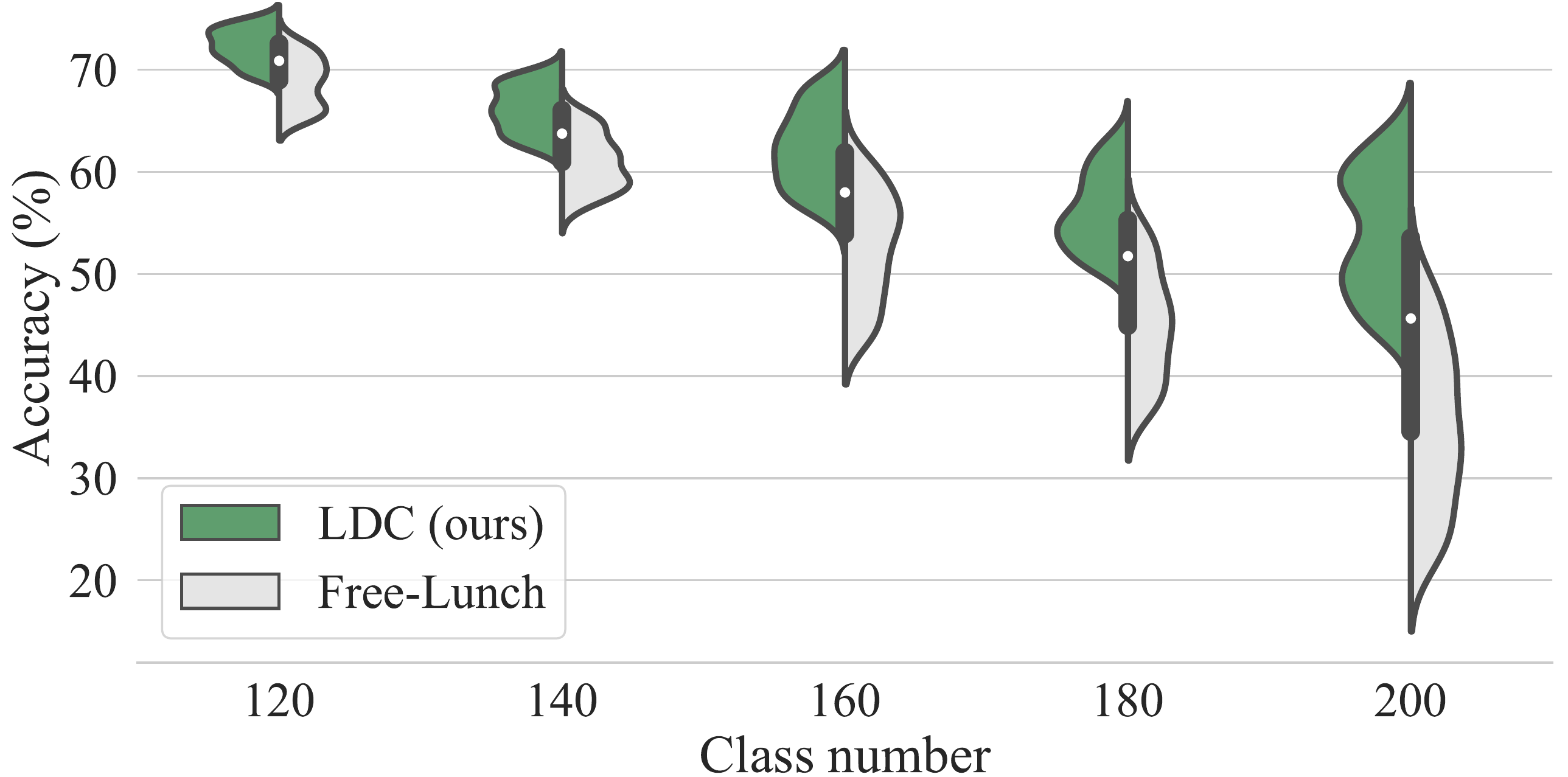} 
\end{center}
\caption{Comparison of Free-Lunch and LDC w.r.t. class numbers.}
\label{fig:violin}
\end{figure}

\begin{table}[t]
\caption{Comparison of covariance matrix mapping strategy (ours) and the strategy adopted by empirical distribution calibration.} 
\label{table:strategy}
\centering
\renewcommand\tabcolsep{4.0pt}
\resizebox{1.0\linewidth}{12mm}{
\begin{tabular}{l|c|cccccccc}
\hline
\hline
\multicolumn{2}{c}{Strategy} &Ours & Empirical \\
\hline
\multirow{2}{*}{mini-ImageNet}&Convergence (epoch) &206 &312 \\
&Generalization (accuracy) & 19.78\% & 15.15\%\\
\hline
\multirow{2}{*}{CIFAR100}&Convergence (epoch) &98 &196 \\
&Generalization (accuracy) & 31.36\% & 27.55\%\\
\hline
\end{tabular}}
\end{table}

\begin{table*}[t]
\caption{Performance comparison on CUB200 using Resnet18. `$\mathrm{PD}$' denotes the performance drop and `$\mathrm{PR}$' the performance retention. `$\ddag$' denotes the method re-implemented under FSCIL settings.}
\label{table:SOTA_CUB}
\begin{center}
\renewcommand\tabcolsep{4.0pt}
\resizebox{1.0\textwidth}{31mm}{
\begin{tabular}{lccccccccccccc}  
\hline
\hline
\multirow{2}{*}{Method} & \multicolumn{11}{c}{Accuracy in each session (\%)} & \multirow{2}{*}{$\mathrm{PD} \downarrow$} &     \multirow{2}{*}{$\mathrm{PR} \uparrow$} \\       
\cline{2-12}
& 0 & 1 & 2 & 3 & 4 & 5 & 6 & 7 & 8 & 9 &10 & & \\      
\Xhline{1.0pt}
Ft-CNN &68.68 &44.81 &32.26&25.83 &25.62 &25.22 &20.84 &16.77 &18.82 &18.25&17.18 &51.50 &25.00\%\\
\hline
iCaRL~\cite{iCaRL2017} &68.68 &52.65 &48.61 &44.16 &36.62 &29.52 &27.83 &26.26 &24.01 &23.89 &21.16 &36.67 &30.80 \%\\
TOPIC~\cite{TOPIC2020}  &68.68 &62.49 &54.81 &49.99 &45.25 &41.40 &38.35 &35.36 &32.22 &28.31 &26.28 &42.40 &38.26 \%\\ 
SPPR~\cite{Self-Promoted2021} &68.68 &61.85 &57.43 &52.68 &50.19 &46.88 &44.65 &43.07 &40.17 &39.63 &37.33 &31.35 &54.35\% \\
FSLL~\cite{FSLL2021} & 72.77 &69.33 &65.51 &62.66 &61.10 &58.65 &57.78 &57.26 &55.59 &55.39 &54.21 &17.38 &74.49\% \\
SFMS~\cite{SynthesizedFeature2021} & 68.78 &59.37 &59.32 &54.96 &52.58 &49.81 &48.09 &46.32 &44.33 &43.43 &43.23 &25.55 &62.85\% \\
CEC~\cite{EvolvedClassifiers2021} &75.85 &71.94 &68.50 &63.50 &62.43 &58.27 &57.73  &55.81 &54.83 &53.52 &52.28 &23.57 &68.92 \%\\ 
Meta-FSCIL~\cite{metafscil} &75.90 &72.41 &68.78 &64.78 &62.96 &59.99 &58.30 &56.85 &54.78 &53.82 &52.64 & 23.26& 69.35\%\\
FACT~\cite{fact} &75.90 &73.23 &70.84 &66.13 &65.56 &62.15 &61.74 &59.83 &58.41 &57.89 &56.94 &18.96 &75.02\%\\
\hline 
Free-Lunch\ddag &77.89 &74.68 &71.36 &66.35 &64.52 &61.63 &58.51 &58.21 &57.31 &56.72 &55.68 &22.25 &71.45$\%$ \\
{\bf {LDC~(ours)}} &\bf77.89 &\bf76.93 &\bf74.64 &\bf70.06 &\bf68.88 &\bf67.15 &\bf64.83 &\bf64.16 &\bf63.03 &\bf62.39 & \bf61.58 &\bf 16.31 &\bf 79.06\% \\ 
\hline
\end{tabular} 
}
\end{center} 
\end{table*} 
\setlength{\tabcolsep}{1.6pt}

\begin{figure*}[t]
\begin{center}
\includegraphics[width=1.0\linewidth]{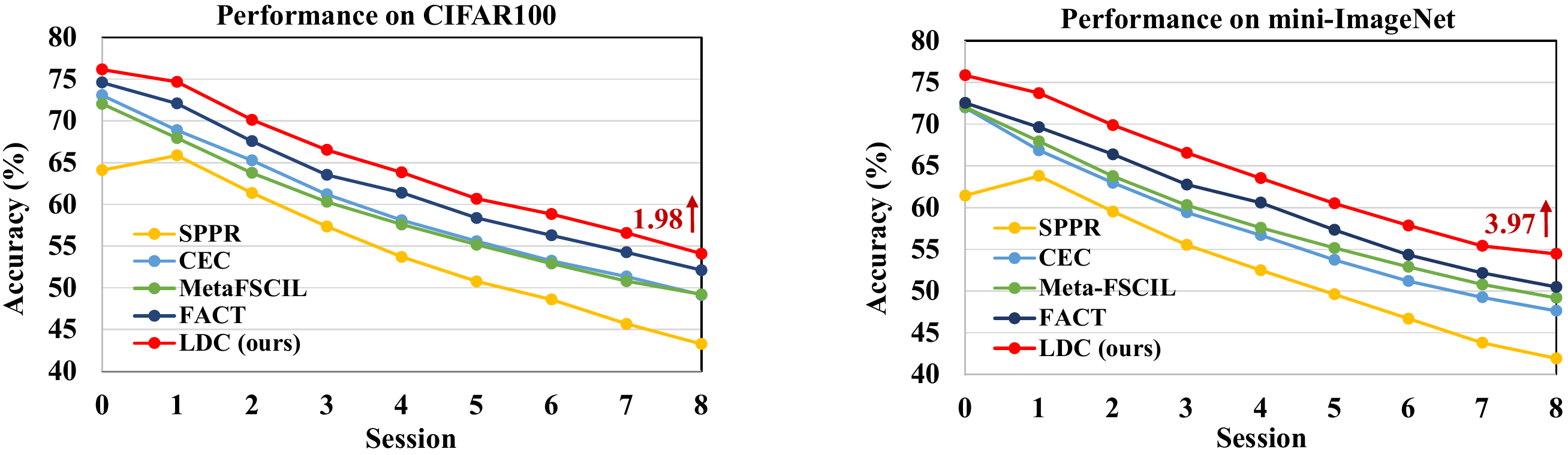} 
\end{center}
\caption{Performance on CIFAR100 and mini-ImageNet datasets.}
\label{fig:line-performance}
\end{figure*}

\textbf{Outlier Cases.} We compare the performance of LDC and Free-Lunch using normal and outlier cases to further validate the ability to calibrate biased distributions, Table~\ref{table:outlier}.
For normal cases, few-shot samples are randomly sampled from new class distributions using the mini-ImageNet dataset. For outlier cases, few-shot samples are outlier samples in new class distributions. The larger performance gains ($i.e.$, $\Delta \mathrm{PD}$ and $\Delta \mathrm{PR}$) in outlier cases demonstrate that LDC achieves better calibration results than Free-Lunch when few-shot samples are extremely biased.

\textbf{Scalability.} We compare performance drops when the class number increases to validate LDC's scalability. We fix the memory costs and calculate the classification accuracy for a fair comparison. From Fig.~\ref{fig:violin}, one can see that when classes increase, LDC maintains the performance while Free-Lunch suffers significant performance drops. The results demonstrate that LDC has higher scalability w.r.t. class number.

\textbf{Covariance matrix mapping strategy.} We compare the convergence and generalization of our proposed covariance matrix mapping strategy and the strategy adopted in empirical distribution calibration~\cite{FreeLunch2021}. As shown in Table~\ref{table:strategy} the model with matrix mapping strategy converges at $206^{th}$ epoch on mini-ImageNet, while the compared model with empirical strategy converges at $312^{th}$ epoch. For generalization, we randomly select 20 unseen classes with few-shot samples for finetuning and test the classification accuracy. The results show that the model with matrix mapping strategy outperforms the counterpart model on both mini-ImageNet and CIFAR100 datasets. The results show that the proposed strategy (sharing mean covariance matrix and mapping the covariance matrix for each class) benefits learning of class distribution statistics, as well as improving the model generalization capability.

\subsection{Performance Comparison}
\textbf{CUB200.} In Table~\ref{table:SOTA_CUB}, LDC achieves the best performance on all sessions and the lowest performance dropping rate. Benefited by PCU, LDC's accuracy in the first session is higher than other methods, and it outperforms FSLL~\cite{FSLL2021} and Fact~\cite{fact} by 7.37\% and 4.64\% in the final session.
\textbf{CIFAR100.} In Fig.~\ref{fig:line-performance}(left), LDC outperforms the state-of-the-arts across all incremental sessions. Specifically, it improves SPPR~\cite{Self-Promoted2021} and Fact~\cite{fact} by 10.83\% and 1.98\%, respectively. \textbf{Mini-ImageNet.} Compared to the baseline method (CEC), LDC improves CEC by 6.83\% (54.46\% v.s. 47.63\%), which demonstrates the effectiveness of the proposed learnable distribution calibration mechanism. Compared to the state-of-the-art methods, LDC outperforms the Fact method by 3.97\%, Fig~\ref{fig:line-performance}(right). 

\begin{table}[t]
\caption{Performance comparison of few-shot classification.} 
\label{table:fsl}
\centering
\renewcommand\tabcolsep{4.0pt}
\resizebox{1.0\linewidth}{13mm}{
\begin{tabular}{l|l|llllll}
\hline
\hline
\multicolumn{2}{c}{Dataset} &Free-Lunch & LDC (ours) \\
\hline
\multirow{2}{*}{mini-ImageNet} & 5-way 1-shot &68.57$\pm$0.55 &\bf69.98$\pm$ 0.42 \\ & 5-way 5-shot & 82.88$\pm$0.42 &\bf84.03$\pm$0.46 \\
\hline
\multirow{2}{*}{CUB} & 5-way 1-shot &79.56$\pm$0.87 &\bf81.51$\pm$0.72 \\ & 5-way 5-shot & 90.67$\pm$0.35 & \bf92.31$\pm$ 0.28 \\
\hline
\end{tabular}}
\end{table}

\subsection{Generalized to Regular Few-Shot Classification}
LDC has great potential to be a plug-and-play module which improves model generalization capacities. To demonstrate such potential, we apply LDC to the regular setting of few-shot classification. The PCU trained using base classes with sufficient training samples is used to generate (augment) samples for novel classes. It improves the accuracy upon Free-Lunch for both 1-shot and 5-shot on mini-ImageNet and CUB datasets, demonstrating the
general effectiveness for few-shot learning problems, Table~\ref{table:fsl}.

\section{Conclusion}    
We proposed a learnable distribution calibration (LDC) mechanism which initiates and estimates all class distributions from a single stored distribution (the covariance matrix). LDC was built upon the parameterized calibration unit (PCU), which reuses network parameters and classifier vectors, thereby has negligible parameter and memory costs. LDC was implemented by driving the PCU to remember class distributions through minimizing distribution matching loss between real and generated samples. LDC provides a fresh insight to solve the catastrophically forgetting and over-fitting problems in a unified framework.

\ifCLASSOPTIONcompsoc

\else
\section*{Acknowledgment}
This work was supported by National Natural Science Foundation of China (NSFC) under Grant 61836012, 61771447 and 62006216.
\fi

\ifCLASSOPTIONcaptionsoff
  \newpage
\fi

\bibliographystyle{ieeetr}
\bibliography{IEEEabrv,egbib}

\end{document}